\documentclass[runningheads]{llncs}
\usepackage{graphicx}
\usepackage{multirow}
\usepackage{arydshln}
\usepackage{booktabs}

\begin{document}
\title{Essay-BR: a Brazilian Corpus of Essays}
%
%\titlerunning{Abbreviated paper title}
% If the paper title is too long for the running head, you can set
% an abbreviated paper title here
%
% \orcidID{0000-1111-2222-3333}

\author{Jeziel C. Marinho\inst{1} \and
Rafael T. Anchiêta\inst{2} \and
Raimundo S. Moura\inst{1}}
\authorrunning{Marinho et al.}
% First names are abbreviated in the running head.
% If there are more than two authors, 'et al.' is used.
%
\institute{Federal University of Piau\'{i}, Brazil \\
\email{\{jezielcm,rsm\}@ufpi.edu.br}\\
\and
Federal Institute of Piau\'{i}, Brazil\\
\email{rta@ifpi.edu.br}}
\maketitle              % typeset the header of the contribution
\begin{abstract}
Automatic Essay Scoring (AES) is defined as the computer technology that evaluates and scores the written 
essays, aiming to provide computational models to grade essays either automatically or with minimal human 
involvement. While there are several AES studies in a variety of languages, few of them are focused on the 
Portuguese language. The main reason is the lack of a corpus with manually graded essays. 
In order to bridge this gap, we create a large corpus with several essays written by Brazilian high school 
students at an online platform. All of the essays are argumentative and were scored across five competencies 
by experts. Moreover, we conducted an experiment on the created	corpus and showed challenges posed by the 
Portuguese language. Our corpus is publicly available at \url{https://github.com/rafaelanchieta/essay}

\keywords{Corpus analysis \and Automatic Essay Scoring \and NLP Education.}
\end{abstract}
\section{Introduction}

The Automated Essay Scoring (AES) area began with Page \cite{page1966imminence} with the Project 
Essay Grader. 
Shermis and Barrera \cite{shermis2002exit} define AES as the computer technology that evaluates and scores 
written essays, i.e., it aims to provide computational models to grade essays either automatically or with minimal 
involvement of humans~\cite{page1966imminence}.

AES is one of the most important educational applications of Natural Language Processing 
(NLP) \cite{ke2019automated,beigman-klebanov-madnani-2020-automated}. It encompasses other fields, such as 
Cognitive Psychology, Education Measurement, Linguistics, and Written Research 
\cite{shermis2013handbook}, which aim to study methods to assist teachers in automatic 
assessments, providing a cheaper, faster, and deterministic approach than humans to grade an essay.

Due to all the benefits, AES has been widely studied in various languages, for example, English, Chinese, 
Danish, Japanese, Norwegian, and Swedish, among others \cite{beigman-klebanov-madnani-2020-automated}.
%the English language through several 
%corpora, as Cambridge-Leaner Corpus-First Certificate in English exam 
%(CLC-FCE) \cite{yannakoudakis-etal-2011-new}, Automated Student Assessment 
%Prize\footnote{\url{https://www.kaggle.com/c/asap-aes}} (ASAP), TOEFL 11 \cite{blanchard2013toefl11}, 
%International Corpus of Learner English (ICLE) \cite{icle}, and 
%Argument Annotated Essays (AAE) \cite{stab-gurevych-2014-annotating}. 
To grade an essay, these studies supported the development of regression-based methods, such as 
\cite{beigman-klebanov-etal-2016-topicality,vajjala2018automated}, classification-based methods, 
as \cite{farra-etal-2015-scoring,nguyen2018argument}, and neural networks-based methods 
as \cite{taghipour-ng-2016-neural}. Moreover, AES systems have also been successfully used in 
schools and large scale exams \cite{williamson2009framework}. 
According to Dikli \cite{dikli2006overview}, examples of such systems are: Intelligent 
Essay\textsuperscript{TM}, Criterion\textsuperscript{SM}, 
IntelliMetric\textsuperscript{TM}, 
E-rater\textsuperscript{\textregistered}, and MY Access!\textsuperscript{\textregistered}.

Despite the importance of AES area, most of the resources and methods are only available for the English 
language \cite{ke2019automated}. There are very few AES-based studies for the Brazilian Portuguese language. 
The main reason for that is the lack of a public corpus with manually graded essays. 
Hence, it is important to put some effort into creating resources that will be useful for the 
development of alternative methods for this field.

In this paper, aiming to fulfill this gap, we create a large corpus, namely Essay-BR, with essays written 
by Brazilian high school students through an online platform. These essays are of the argumentative type 
and were graded by experts across five different competencies to achieve the total score of an essay. 
The competencies follow the evaluation criteria of the ENEM - 
\textit{\textbf{E}xame \textbf{N}acional do \textbf{E}nsino \textbf{M}\'{e}dio} - (National High School Exam), 
which is the main Brazilian high school exam that serves as an admission test for most universities in Brazil. 
%In summary, reviewers evaluated each competence taking into account some proficiency levels, and the sum of 
%the scores of each competence is the holistic 
%score. Thus, the holistic score summarizes the quality of an essay with a single number, whereas the 
%competencies provide better feedback and a fine-grained evaluation of the quality of an essay.
%At this time, the corpus has 3,804 
%essays and 163 prompts. 
%Concerning the popular English corpora, our corpus is the second in the number of 
%essays, while the ASAP corpus is the first with 17,450 essays \cite{ke2019automated}.

%Another aspect of the Essay-BR corpus is its scoring strategies. It has scores for five competencies that are 
%evaluated according to some proficiency levels and a holistic score that is the sum of the five competencies. 

In addition to the corpus, we carry out an experiment, implementing two approaches to automatically score 
essays, demonstrating the challenges posed by the corpus, and providing baseline results. 
As this is the first publicly available corpus of essays for the Portuguese language, we believe that it 
will foster AES studies for that language, resulting in developing alternative methods 
to grade an essay.

The remaining of this paper is organized as follows. Sect.~\ref{sec:related} describes the main related 
work for the Portuguese language. In Sect. \ref{sec:enem}, we present the ENEM exam. 
Sect. \ref{sec:corpus} details our corpus, its construction, and an analysis of the training, development, 
and testing datasets. In Sect. \ref{sec:experiment}, we describe our conducted experiments. 
Finally, Sect.~\ref{sec:conclusion} concludes the paper, indicating future work.

\section{Related Work} \label{sec:related}

As before mentioned, there is no publicly available corpus of essays for the Brazilian Portuguese. 
However, three efforts investigated AES for that language. Here, we briefly present them.

Bazelato and Amorim \cite{bazelato2013bayesian} crawled $429$ graded essays from the \textit{Educação UOL} Website 
to create the first corpus of essays for the Portuguese Language. However, the crawled essays are very old, 
they do not meet the criteria of the ENEM exam. Moreover, the collected essays are not available.

Amorim and Veloso \cite{amorim2017multi} developed an automatic essay scoring method for the Brazilian Portuguese 
language. For that, they collected $1,840$ graded essays from the \textit{Educação UOL} Website. Next, they 
developed 19 features to feed a linear regression to grade the essays. Then, to evaluate the approach, 
the authors compared the automatic scores with the scores of the essays, using the Quadratic Weighted Kappa 
(QWK) metric~\cite{cohen1968weighted}, achieving 42.45\%.

Fonseca et al. \cite{fonseca2018automatically} addressed the task of automatic essay scoring in two ways. 
In the first one, they adopted a deep neural network architecture similar to the Dong et al. \cite{dong2017attention} 
with two Bidirectional Long Short-Term Memory (BiLSTM) layers. The first layer reads word vectors and 
generates sentence vectors, which are read by the second layer to produce a single essay vector. 
This essay vector goes through an output layer with five units and a sigmoid activation function to get an 
essay score. In the second approach, the authors hand-crafted 681 features to feed a regressor to grade an 
essay. The authors evaluated the approaches using a corpus with 56,644 graded essays and reached the best 
result with the second method, achieving 75.20\% in the QWK metric.

Although these works have used essays written in Brazilian Portuguese to evaluate their methods, the authors 
did not make corpora publicly available, making the development of alternative methods difficult. 
Moreover, each work utilized a different corpus, also making it difficult to provide a fair comparison between them.

In English, according to \cite{ke2019automated}, there are five popular available corpora: 
ICLE~\cite{icle}, CLC-FCE \cite{yannakoudakis-etal-2011-new}, Automated Student Assessment 
Prize\footnote{\url{https://www.kaggle.com/c/asap-aes}} (ASAP), TOEFL 11 \cite{blanchard2013toefl11}, 
and AAE~\cite{stab-gurevych-2014-annotating}. The ASAP corpus, one of the 
most famous and established corpus, was released as part of a Kaggle competition in 2012, becoming widely 
used for holistic scoring.Furthermore, the corpus is composed by 17,450 argumentative essays and 8 prompts 
written by United States students from grades 7 to 10.

In what follows, we present the ENEM exam.

\section{ENEM exam} \label{sec:enem}

The ENEM - \textit{\textbf{E}xame \textbf{N}acional do \textbf{E}nsino \textbf{M}\'{e}dio} - (National High School Exam) 
is the main Brazilian high school exam that serves as an admission test for most 
universities in Brazil. More than that, it is the second-largest admission test in the world after the National 
Higher Education Entrance Examination, the entrance examination of higher education in China. 
In the ENEM exam, the reviewers take into account five competencies to evaluate an essay, which are:

\begin{enumerate}
	\item Adherence to the formal written norm of Portuguese.
	\item Conform to the argumentative text genre and the proposed topic (prompt), to develop a text, 
	using knowledge from different areas.
	\item Select, relate, organize, and interpret data and arguments in defense of a point of view.
	\item Usage of argumentative linguistic structures.
	\item Elaborate a proposal to solve the problem in question.
\end{enumerate}

\noindent where each competence is graded with scores ranging from 0 to 200 in intervals of 40. 
These scores are organized by proficiency levels, as shown in Table \ref{tab:levels}. In this table, 
the value 200 indicates an excellent proficiency in the field of competence, whereas the value 0 shows 
ignorance in the field of competence.

In this way, the total score of an essay is the sum of the competence scores and may range from 0 to 1,000. 
At least two reviewers grade an essay in the ENEM exam, and the final score for each competence is the 
arithmetic mean between the two reviewers. If the disagreement between the reviewers' score is greater 
than 80, a new reviewer is invited to grade the essay. Thus, the final score for each competence will be 
the arithmetic mean between the three reviewers.

\begin{table}[ht]
	\centering
	\caption{Proficiency levels of the ENEM exam} \label{tab:levels}
	\begin{tabular}{cp{5.9cm}}
		\toprule
		\textbf{Score} & \textbf{Description} \\
		\midrule
		200 & excellent proficiency \\
		160 & good mastery \\
		120 & medium dominance \\
		80	& insufficient mastery \\
		40	& precarious dominance \\
		0	& ignorance \\
		\bottomrule
	\end{tabular}
\end{table}

\section{Essay-BR Corpus} \label{sec:corpus}

The Essay-BR corpus contains $4,570$ argumentative documents and $319$ topics (prompts). They were 
collected from December 2015 to April 2020. The topics include: human rights, political issues, 
healthcare, cultural activities, fake news, popular movements, covid-19, and others. 
Also, they are annotated with scores in the five competencies of the ENEM exam. 
Table~\ref{tab:essay_details} summarizes the Essay-BR corpus.

\begin{table}[ht]
	\centering
	\caption{Summary of the Essay-BR Corpus} \label{tab:essay_details}
	\begin{tabular}{cc}
		\toprule
		\textbf{Details} & \textbf{Essay-BR} \\
		\midrule
		Text type & Argumentative \\
		Writer's language level & BR students (high school) \\
		Scoring & Holistic \\
		Number of essays & $4,570$ \\
		Number of prompts & $319$ \\
		Number of competencies & $5$ \\
		proficiency range & $[0 - 200]$ \\
		proficiency scores & $0, 40, 80, 120, 160, 200$ \\
		Score range & $[0 - 1,000]$ \\
		\bottomrule
	\end{tabular}
\end{table}

\subsection{Construction of the Essay-BR corpus}

To create the Essay-BR corpus, we developed a Web Scraper to extract essays from two public Websites: 
\textit{Vestibular UOL}\footnote{\url{https://vestibular.brasilescola.uol.com.br/banco-de-redacoes}} and 
\textit{Educa\c{c}\~{a}o UOL}\footnote{\url{https://educacao.uol.com.br/bancoderedacoes}}. The essays from 
these Websites are public, may be used for research purposes, were written by high school students, 
and are graded by experts following the ENEM exam criteria. We collected $798$ essays and $43$ 
prompts from \textit{Educação UOL}, and $3,772$ essays and $276$ prompts from \textit{Vestibular UOL}. 
The difference in the number of essays is because of the latter Website receives up to forty essays per 
month, while the former receive up to twenty essays per month.

\begin{figure}[ht]
	\centering
	\includegraphics[scale=0.4]{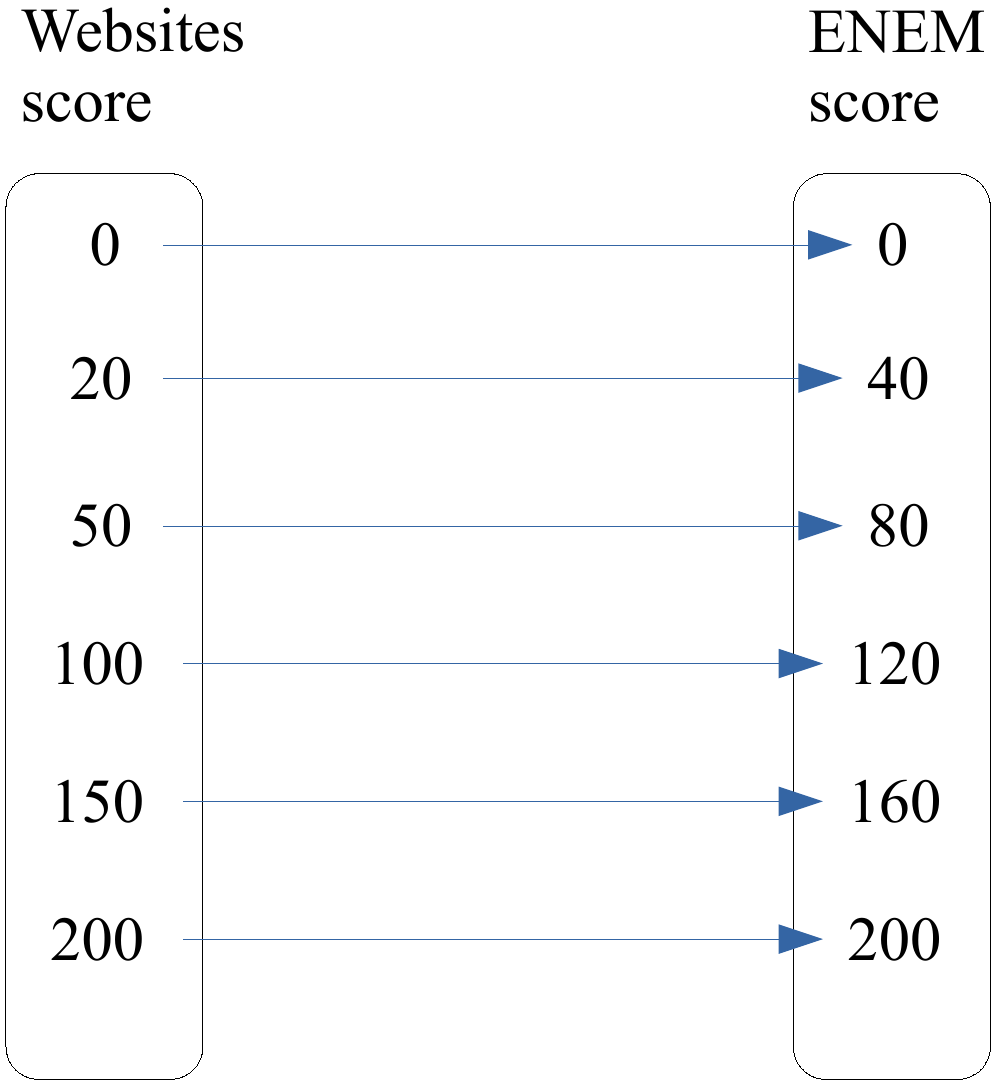}
	\caption{Mapping function from Website scores to ENEM scores} \label{fig:mapping}
\end{figure}

After collecting the essays, we applied some preprocessing to remove HTML tags and comments 
from the reviews. So, the essays contain only the content written by the students. Then, we normalized the 
scores of the essays. Although these Websites adopted the same ENEM exam competencies to evaluate 
the essays, they have a slightly different scoring strategy. Thus, we mapped the scores from Websites to ENEM 
scores, as shown in Figure \ref{fig:mapping}.

% As we can see, we mapped the competence scores from Websites to ENEM score.
%Following the above steps, we collected 3,804 essays about 163 topics (prompts) from December 2015 to 
%April 2020. These topics include: human rights, political issues, healthcare, cultural activities, fake 
%news, popular movements, covid-19, among others. Table~\ref{tab:essay_details} summarizes the Essay-BR corpus.

%\subsection{Aspects of the Essay-BR corpus}

%The Essay-BR is a corpus with argumentative type essays. The essays were written by Brazilian high school 
%students on an online platform and graded by human professionals concerning the five competencies of the 
%ENEM exam. 

Although the corpus has a holistic scoring, it also has proficiency scores. 
Holistic scoring technologies are commercially valuable, since they allow automatically scoring million 
of essays deterministically, summarizing the quality of an essay with a single score. 
However, it is not adequate in classroom settings, where providing students with feedback on how to 
improve their essays is of utmost importance \cite{ke2019automated}. To mitigate this weakness, the 
Essay-BR corpus contains five competencies. Thus, the score of competence shows how a student should 
improve its essay. For example, a student who got a score equals 40 in the first competence, i.e., 
adherence to the formal written norm, got as feedback that it is necessary to improve its grammar.

We also present an example of the structure of our corpus, as shown in Table~\ref{tab:corpus}. 
From this table, the score is the sum of the competencies (C1 to C5), and the essay content is 
composed as a list of paragraphs. It is important to say that some essays have no title, since, 
in the ENEM exam, the title is not mandatory.

\begin{table}[ht]
	\centering
	\caption{Example of the Essay-BR corpus} \label{tab:corpus}
	\begin{tabular}{cc}
		\toprule
		\textbf{Attribute} & \textbf{Value} \\
		\midrule
		Prompt & covid-19 \\
		Score & 720 \\
		Title & Fighting coronavirus through science \\
		Essay content & \textit{list of paragraphs} \\
		C1 & 160 \\
		C2 & 160 \\
		C3 & 120 \\
		C4 & 160 \\
		C5 & 120 \\
		\bottomrule
	\end{tabular}
\end{table}

Besides the structure, we computed some statistics and linguistics features about the 
essays of the corpus, as depicted in Tables~\ref{tab:statistics} and \ref{tab:linguistics}, 
respectively.
%In order to analysis the essays of the corpus, we computed some statistics about the them. 
%In , we present the calculated statistics. 
In Table \ref{tab:statistics}, we can see that, on average, an essay has 4 paragraphs, and 
each paragraph has 2 sentences. Furthermore, the sentences are somewhat long, with an 
average of 30 tokens. In Table \ref{tab:linguistics}, most of the essays are passive voice, 
since in the Portuguese Language the essays should be impersonal. Also, we calculated the 
Flesch score that measures the readability of an essay. From the score value, the essays 
are compatible with the college school level. Finally, we computed some richness vocabulary 
metrics, such as: hapax legomenon, which is a word that occurs only once, lexical diversity, 
also known as the type-token ratio, and lexical density, which is the number of lexical tokens 
divided by the number of all tokens.

\begin{table}[ht]
	\centering
	\begin{minipage}[b]{0.49\textwidth}
	    \caption{Statistics of the essays} \label{tab:statistics}
	    \begin{tabular}{ccc}
		    \toprule
		    \textbf{Statistic} & \textbf{Mean} & \textbf{Std} \\
		    \midrule
		    Paragraph per essay & 4.08 & 1.15 \\
		    Sentence per essay & 10.57 & 4.42 \\
		    Sentence per paragraph & 2.58 & 1.44 \\
		    Token per essay & 324.40 & 94.19 \\
		    Token per paragraph & 79.33 & 35.22 \\
		    Token per sentence & 30.66 & 17.68 \\
		    \bottomrule
	    \end{tabular}
	\end{minipage}
	\hfill
	\begin{minipage}[b]{0.39\textwidth}
	    \caption{Linguistics features} \label{tab:linguistics}
	    \begin{tabular}{cc}
	         \toprule
	         \textbf{Feature} & \textbf{Mean} \\
	         \midrule
	         Passive voice & 75\% \\
	         Active voice & 25\% \\
	         lexical diversity & 26\% \\
	         lexical density & 22\% \\
	         Flesch score & 45.98 \\
	         Hapax legomenon & 36.46 \\
	         \bottomrule
	    \end{tabular}
	\end{minipage}
\end{table}

\begin{figure}[ht]
	\centering
	\includegraphics[scale=0.45]{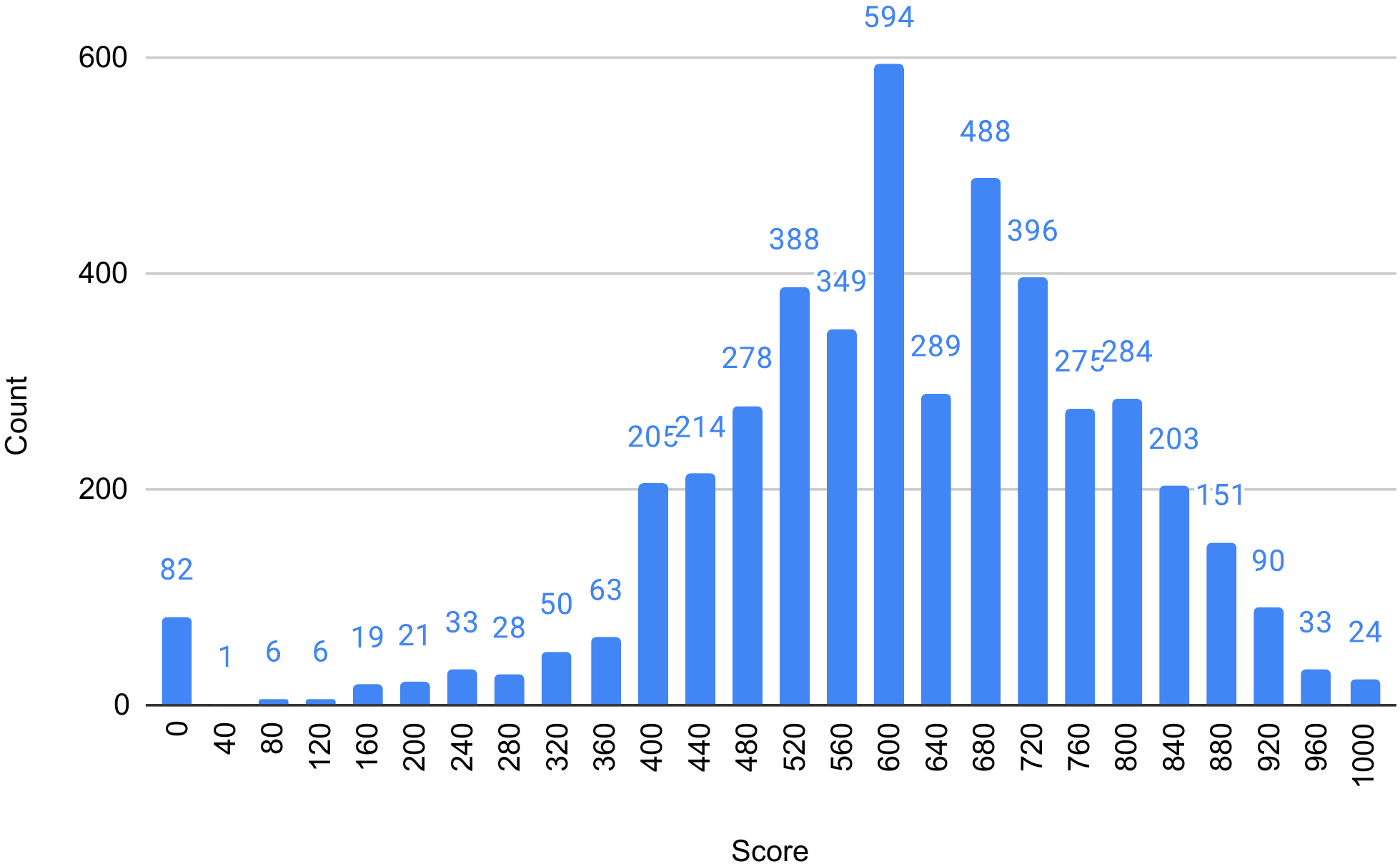}
	\caption{Distribution of the total score} \label{fig:distribution}
\end{figure}

In order to create sets of training, development, and testing, we divided the corpus in proportions of 
$70\%$, $15\%$, and $15\%$, which corresponds to $3,198$, $686$, and $686$ essays for training, development, and 
testing, respectively. Aiming to choose essays with a fair distribution of scores for each split, we 
computed the distribution of the total score of the essays, as depicted in Figure \ref{fig:distribution}.

The top $3$ scores are $600$, $680$, and $720$ corresponding to $13.00\%$, $10.68\%$, and $8.67\%$ of 
the corpus, respectively, indicating that essays with these scores should appear more times in the 
training, development, and testing sets. Moreover, the scores in the corpus have a slightly rightward 
skewed normal distribution.

We also computed the distribution score for each competence and presented it in 
Table \ref{tab:competencies}. From this table, all of the essays received $120$ as the 
higher score, showing that, in general, the students have medium dominance in the field 
of competence. 
%However, in the third competence, that analyzes the competence of a student in selecting, relating, 
%organizing, and interpreting data and arguments in defense of a point of view, most of the students 
%received score equals 80, showing an insufficient dominance in this competence.

\begin{table}[ht]
	\centering
	\caption{Distribution score for each competence} \label{tab:competencies}
	\begin{tabular}{ccccccc}
		\toprule
		\multirow{2}{*}{\textbf{Competence}} & \multicolumn{6}{c}{\textbf{Scores}} \\
		& 0 & 40 & 80 & 120 & 160 & 200 \\
		\midrule
		C1 & 97 & 24 & 359 & \textbf{2,630} & 1,338 & 122 \\
		C2 & 109 & 79 & 689 & \textbf{1,711} & 1,705 & 277 \\
		C3 & 122 & 146 & 1,206 & \textbf{2,130} & 827 & 139 \\
		C4 & 134 & 61 & 590 & \textbf{2,000} & 1,241 & 544 \\
		C5 & 302 & 276 & 1,023 & \textbf{1,732} & 815 & 422 \\
		\bottomrule
	\end{tabular}
\end{table}

In the following subsection, we analyzed the training, development, and testing sets of the Essay-BR 
corpus.

\subsection{Analysis of the Essay-BR corpus} \label{sec:analysis}

\begin{figure}[ht]
    \centering
    \includegraphics[scale=0.45]{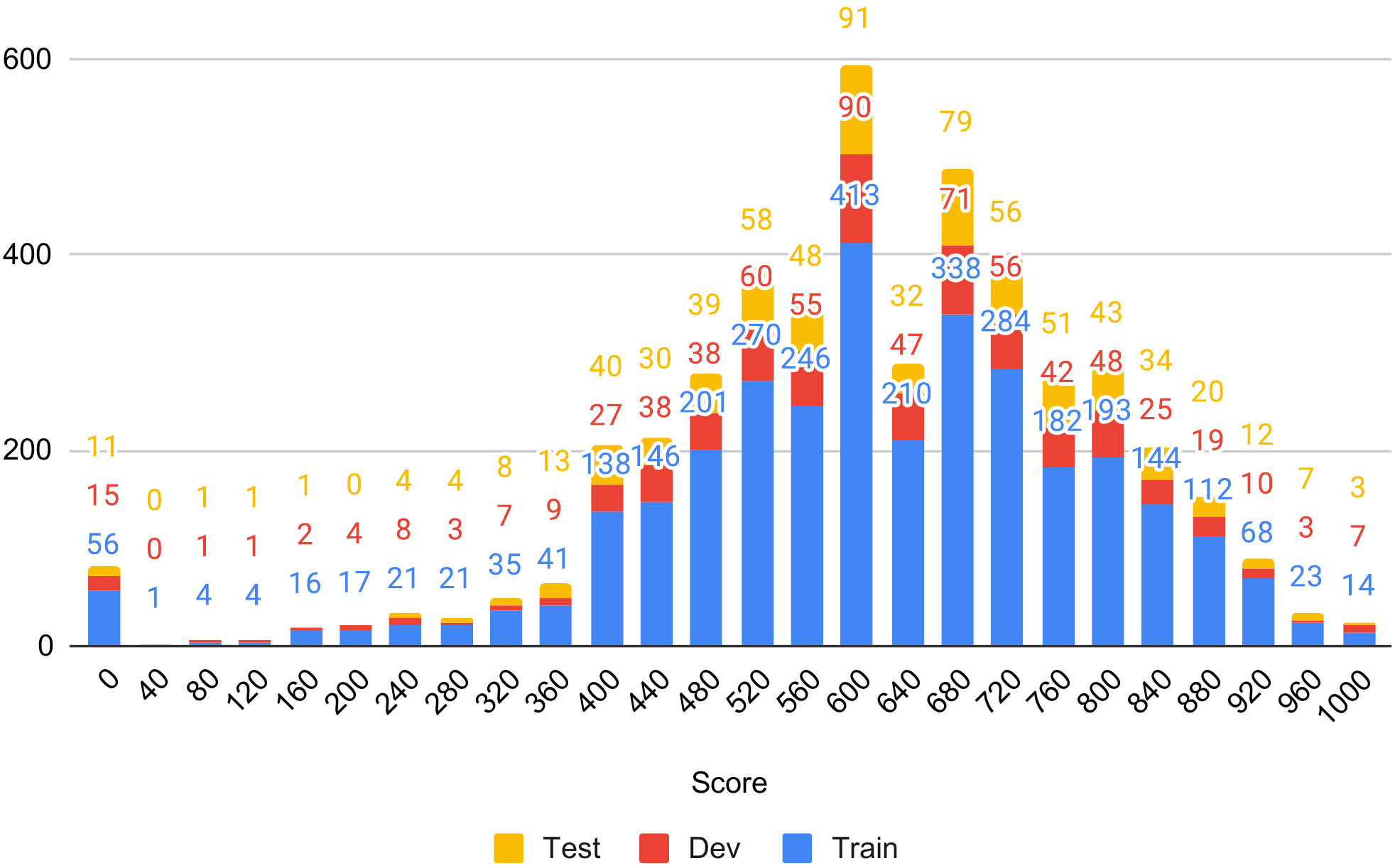}
    \caption{Training, development, and testing sets of the Essay-BR corpus}
    \label{fig:stacked}
\end{figure}

%As mentioned, we split the corpus in 80\%, 10\%, and 10\% for training, development, and testing, which 
%corresponding to 3,043, 380, and 381 essays, respectively. 
To create the three splits with score distributions similar to that of the complete corpus, we first 
shuffled all the data; then, we filled each split with essays based on the score distribution. 
Figure \ref{fig:stacked} presents the score distribution for the 
training, development, and testing sets, respectively.

%\begin{figure}[ht]
%	\centering
%	\includegraphics[scale=0.38, angle=90]{img/train2.pdf}
%	\caption{Training set of the Essay-BR corpus} \label{fig:train}
%\end{figure}

%\begin{figure}[ht]
%	\centering
%	\includegraphics[scale=0.38]{img/dev.pdf}
%	\caption{Development set of the Essay-BR corpus} \label{fig:dev}
%\end{figure}

%\begin{figure}[ht]
%	\centering
%	\includegraphics[scale=0.38]{img/test.pdf}
%	\caption{Testing set of the Essay-BR corpus} \label{fig:test}
%\end{figure}

From this figure, one can see that the score distributions are similar to the score distribution of the 
complete corpus. Likewise, in the score 
distribution of Figure \ref{fig:distribution}, the top $3$ scores of the training set are $600$, $680$, and $720$. 
Moreover, the development and testing sets have a similar distribution. This means that if a machine-learning 
algorithm performs well in the development set, it probably will also perform well in the testing set.

More than the scores, we also calculated some statistics on the splits, intending to 
verify whether the proportion of paragraphs, sentences, and tokens for each division remained related to 
the complete corpus proportion.

Comparing the obtained results in Table \ref{tab:competencies} with the got results of each split in 
Table \ref{tab:stats_splits}, we can see that the results maintained similar proportions. For example, 
the average of paragraphs per essay, sentences per essay, and sentences per paragraph had related 
results: $4$, $10$, and $2$, respectively.

\begin{table}[ht]
	\centering
	\caption{Statistics for each split of the corpus} \label{tab:stats_splits}
		\begin{tabular}{cccc}
			\toprule
			\textbf{Split} & \textbf{Statistic} & \textbf{Mean} & \textbf{Standard deviation} \\
			\midrule
			\multirow{6}{*}{Train} & Paragraph per essay & 4.08 & 1.08 \\
			& Sentence per essay & 10.60 & 4.45 \\
			& Sentence per paragraph & 2.59 & 1.44 \\
			& Token per essay & 325.01 & 94.38 \\
			& Token per paragraph & 79.52 & 35.00 \\
			& Token per sentence & 30.64 & 17.70 \\
			\hdashline
			\multirow{6}{*}{Dev} & Paragraph per essay & 4.15 & 1.60 \\
			& Sentence per essay & 10.57 & 4.63 \\
			& Sentence per paragraph & 2.54 & 1.43 \\
			& Token per essay & 323.52 & 95.69 \\
			& Token per paragraph & 77.84 & 35.65 \\
			& Token per sentence & 30.60 & 17.75 \\
			\hdashline
			\multirow{6}{*}{Test} & Paragraph per essay & 4.03 & 0.93 \\
			& Sentence per essay & 10.45 & 4.10 \\
			& Sentence per paragraph & 2.59 & 1.43 \\
			& Token per essay & 322.48 & 91.75 \\
			& Token per paragraph & 79.95 & 35.78 \\
			& Token per sentence & 30.83 & 17.53 \\
			\bottomrule
	\end{tabular}
\end{table}

In what follows, we present the experiment and obtained results.

\section{Experiments and Results} \label{sec:experiment}

We carried out an experiment on the Essay-BR corpus to understand the challenges introduced by the corpus. 
For that, we implemented the feature-based methods of Amorim and Veloso \cite{amorim2017multi} and 
Fonseca et al. \cite{fonseca2018automatically}. We are aware that, in recent years, the NLP area has 
been dominated by the transformer architectures, as BERT \cite{devlin-etal-2019-bert}. 
However, for the AES field the obtained results by these architectures are similar to traditional models, 
such as \textit{N}-grams at high computation cost \cite{mayfield-black-2020-fine}. 
Thus, as a baseline, we preferred to implement feature-based methods since they require less computational 
resources and effort.

Amorim and Veloso \cite{amorim2017multi} developed 19 features: number of grammatical errors, number of verbs, 
number of pronouns, and others. These features fed a linear regression to score an essay. 
Fonseca et al. \cite{fonseca2018automatically} created a pool of 681 features, as the number of discursive markers, 
number of oralities, number of correct words, among others, and these features fed the gradient boosting 
regressor to score an essay. To extract the Essay-BR corpus features, we used the same tools 
reported by the authors, and to implement the regressors, we used the scikit-learn library~\cite{scikit-learn}.

We evaluated those methods using the Quadratic Weighted Kappa (QWK), which is a metric commonly used to 
assess AES models \cite{yannakoudakis-cummins-2015-evaluating}, and the Root Mean Squared Error 
(RMSE), which is a metric employed to regression problems. Table \ref{tab:qwk_results} shows the 
QWK metric results, while Table \ref{tab:rmse_results} presents the results for the RMSE metric. 
In the QWK metric, the greater the value, the better the result, whereas in the RMSE metric, the smaller the 
value, the better the result. 

Although the approach of Fonseca et al. \cite{fonseca2018automatically} achieved better results in both metrics for each 
competence (C1 to C5), these results are not fit for summative student assessment, as normally for the AES field, 
threshold values between $0.6$ and $0.8$ QWK are used as a floor for testing purposes 
\cite{mayfield-black-2020-fine}. Furthermore, the method of Fonseca et al. \cite{fonseca2018automatically}, which achieved 
$75.20\%$ in the QWK metric in their corpus, reached only $51\%$ in the Essay-BR. This difference may be due to 
two factors. The first is the size of the corpus. Fonseca et al. \cite{fonseca2018automatically} used a corpus 
with more than $50,000$ essays, whereas our corpus has $4.570$ essays. The second is implementation details. 
Fonseca et al. \cite{fonseca2018automatically} used several lexical resources, but they did not make them available. Thus, 
we do not know if the lexical resources we used are the same as Fonseca et al.~\cite{fonseca2018automatically}.
%, showing that our corpus is more challenging than that corpus.

As we can see, it is necessary to develop more robust methods to grade essays for the Portuguese language 
in order to improve the results.

\begin{table}[ht]
	\centering
	\caption{Quadratic Weighted Kappa on the test set} \label{tab:qwk_results}
		\begin{tabular}{ccccccc}
			\toprule
			\textbf{Model} & \textbf{C1} & \textbf{C2} & \textbf{C3} & \textbf{C4} & \textbf{C5} & \textbf{Total} \\
			\midrule
			Amorim and Veloso \cite{amorim2017multi} & 0.35 & 0.44 & 0.39 & 0.37 & 0.34 & 0.47 \\
			Fonseca et al. \cite{fonseca2018automatically} & 0.42 & 0.46 & 0.40 & 0.45 & 0.36 & 0.51 \\
			\bottomrule
		\end{tabular}
\end{table}

\begin{table}[ht]
	\centering
	\caption{Rooted Mean Squared Error on the test set} \label{tab:rmse_results}
		\begin{tabular}{ccccccc}
			\toprule
			\textbf{Model} & \textbf{C1} & \textbf{C2} & \textbf{C3} & \textbf{C4} & \textbf{C5} & \textbf{Total} \\
			\midrule
			Amorim and Veloso \cite{amorim2017multi} & 34.82 & 36.14 & 40.27 & 42.23 & 49.02 & 163.92 \\
			Fonseca et al. \cite{fonseca2018automatically} & 34.16 & 35.76 & 40.02 & 40.48 & 48.28 & 159.44 \\
			\bottomrule
		\end{tabular}
\end{table}

% \url{http://github.com/<omitted_due_to_blind_review>}

% \subsection{Utility of the Essay-BR corpus}

%The purpose of the Essay-BR corpus is to allow the development of methods to automatic essay scoring both 
%for holistic scoring and a fine-grained scoring, through the proficiency levels. More than these, 
%according to \citet{ke2019automated}, the quality of an essay may be graded adopting different 
%dimensions, as presented in Table \ref{tab:e_quality}.

\section{Final remarks} \label{sec:conclusion}

% annotate with argument relations

In this paper, we presented a large corpus of essays written by Brazilian high school students that 
were graded by experts following the evaluation criteria of the ENEM exam. This is the 
first publicly available corpus for that language. At this time, it has $4,570$ essays and $319$ prompts, 
but we already scraped 13,306 essays from the \textit{Vestibular UOL} Website. These essays are being 
pre-processed and will be available as soon as possible. We hope that this resource will foster the 
research area for Portuguese by developing of alternative methods to grade essays. 
More than that, according to Ke and Ng \cite{ke2019automated}, the quality of an essay may be graded adopting 
different dimensions, as presented in Table \ref{tab:e_quality}.

\begin{table}[th]
	\centering
	\caption{Dimensions to grade the quality of an essay} \label{tab:e_quality}
		\begin{tabular}{cc}
			\toprule
			\textbf{Dimension} & \textbf{Description} \\
			\midrule
			Grammaticality & Grammar analysis \\
			Usage & Use of prepositions, word usage \\
			Mechanics & Spelling, punctuation, capitalization \\
			Style & Word choice, sentence structure variety \\
			Relevance & Relevance of the content to the prompt \\
			Organization & How well the essay is structured \\
			Development & Development of ideas with examples \\
			Cohesion & Appropriate use of transition phrases \\
			Coherence & Appropriate transitions between ideas \\
			Thesis clarity & Clarity of the thesis \\
			Persuasiveness & Convincingness of the major argument \\
			\bottomrule
		\end{tabular}
\end{table}

From this table, one can see that a corpus of essays may be graded regarding several dimensions. 
Assessing and scoring these dimensions helps the students get better feedback on their essays, 
supporting them to identify which aspects of the essay need improvements.

Some of these dimensions do not seem challenging, such as the grammaticality, usage and mechanism dimensions, 
since they already have been extensively explored. Several other dimensions, such as cohesion, coherence, 
thesis clarity, and persuasiveness, bring problems that involve computational modeling in different levels of the text. 
Modeling these challenging dimensions may require understanding the essay content and exploring the semantic 
and discourse levels of knowledge. Thus, there exist several possible applications that the Essay-BR may be 
useful.

As future work, besides increasing the corpus, which is already in process, we intend to provide the essay 
corrections, aiming to develop machine learning models to learn from the corrections.

%\section*{Acknowledgments}

%The authors are grateful to Federal University of Piauí, Federal Institute of Maranhão, and Federal Institute 
%of Piauí for supporting this work.
%Do not include this section when submitting your paper for review.
%
% ---- Bibliography ----
%
% BibTeX users should specify bibliography style 'splncs04'.
% References will then be sorted and formatted in the correct style.
%
\bibliographystyle{splncs04}
\bibliography{reference}
\end{document}